\documentclass[letterpaper, 10 pt, conference]{ieeeconf}  

\IEEEoverridecommandlockouts                              

\usepackage{graphics} 
\usepackage{epsfig} 
\usepackage{amsmath} 
\usepackage{amssymb}  
\usepackage{algpseudocode}
\usepackage{subfigure}
\usepackage{color}
\usepackage{verbatim}

\title{\LARGE \bf
Description and Technical specification of Cybernetic Transportation Systems: an urban transportation concept
\begin{center}
\textit{\small-CityMobil2 project approach-}
\end{center}
}

\author{Luis Rold\~ao, Joshue P\'erez, David Gonz\'alez and  and Vicente Milan\'es
\thanks{The authors are with the INRIA / Imara team, National Institute for Research in Computer Science and Control (INRIA), Domaine de Voluceau, BP 105, 78153 Le Chesnay, France {\tt\small {luis.roldao, david.gonzalez, joshue.perez\_rastelli, vicente.milanes}@inria.fr}}%
}

\begin{document}

\maketitle
\thispagestyle{empty}
\pagestyle{empty}


\begin{abstract}
The Cybernetic Transportation Systems (CTS) is an urban mobility concept based on two ideas: the car sharing and the automation of dedicated systems with door-to-door capabilities. In the last decade, many European projects have been developed in this context, where some of the most important are: Cybercars, Cybercars2, CyberMove, CyberC3 and CityMobil. Different companies have developed a first fleet of CTSs in collaboration with research centers around Europe, Asia and America. Considering these previous works, the FP7 project CityMobil2 is on progress since 2012. Its goal is to solve some of the limitations found so far, including the definition of the legal framework for autonomous vehicles on urban environment. This work describes the different improvements, adaptation and instrumentation of the CTS prototypes involved in European cities. Results show tests in our facilities at INRIA-Rocquencourt (France) and the first showcase at Leo\'n (Spain). 
\end{abstract}


\section{Introduction}

In the last decade, some European initiatives have been carried out in the framework of Cybernetic Transportation Systems (CTSs), where some of the most important are: Cybercars, Cybercars2 \cite{Naranjo2009}, CyberMove, CityMobil \cite{Bouraoui2006} and CATS. A good overview of the past technologies and the present developments of the CTSs advances is given in \cite{Gonzalez2013}. CTSs were defined in the first Cybercar project as follows: ``CTSs are road vehicles with fully automated driving capabilities, where a fleet of such vehicles forms a transportation system, for passengers or goods, on a network of roads with on-demand and door-to-door capability'' \cite{CyberCars2002}.

In this context, CTSs use different perception and communication technologies for the recognition of the environment. This information is processed for the motion planning and onboard control, in order to guarantee the performance of the vehicles in dynamic scenarios (considering other agents involved on the trajectory). This system proposed different modules to evaluate the feasibility for its behavior in Asian cities, including a central control room, stations, road monitoring, etc \cite{Xia2010}.

Within the CTS framework, recent works are related to autonomous maneuvers and motion planning in urban areas. Inevitable Collision State (ICS) for obstacle avoidance, no matter what the future trajectory of the robotic system, has been implemented in \cite{Martinez-Gomez2009}. On the other hand, autonomous navigation of Cybercars based on Geographical Information Systems (GIS) were described in Bonnifait et al. \cite{Bonnifait2008}. Furthermore, the improvements for the localization in urban environment have been done by the implementation of Kalman filters considering kinematics behavior \cite{Rong2008}, using GPS and Inertial Measurement Units (IMU) and lidar-based lane marker detection using Bayesian and map matching algorithms.

CTSs are also capable to interact with other autonomous and semi-autonomous vehicles, as was demonstrated in \cite{Naranjo2009}. Nevertheless, in most of these scenarios, the trajectory path was previously recorded. Recently, new kinds of scenarios, as signalized and unsignalized intersections, roundabouts and highly risk collisions situations have been considered \cite{Premebida2006, Perez2011D}. Latest development from RITS team at INRIA\footnote{https://team.inria.fr/rits/} are focused on new algorithm generation for, new algorithms for dynamic path generation in urban environments, taking into account structural and sudden changes in straight and bend segments has been recently developed in a simulated environment \cite{Gonzalez2013, Perez2014}.

This paper presents the integration of these previous algorithms \cite{Gonzalez2013, Perez2014} in real scenarios, as part of the CityMobil2 project. A description of the control architecture, in terms of software, hardware and technical/functional specifications of the Cybercars involved in the showcases of the project are also described. The main contribution of this work is the implementation of a dynamic trajectory generation, with scheduled and emergency stops, considering other vehicles and pedestrians in real scenarios with CTSs.

The rest of the paper is organized as follow: a review of the general architecture used for CTSs is explained in Section II. Details of the implementation of the global and local planning are also given. Section III explains the CTSs used at INRIA for demonstrations in urban environments. Tests and results validation is described in Section IV. These were tested in our facilities at INRIA (France) and the first showcase of the CityMobil2 project at Leon (Spain). Finally, conclusions and the future works are listed in Section V.


\section{General architecture} 

In the literature, different approaches for control architectures in autonomous vehicles have been presented. Most of them keep a common structure. Autonomous vehicle demonstrations as performed by: Stanley \cite{thrun2006stanley} -winner of the Darpa Urban Challenge-, the Boss \cite{Ferguson2008} -winner of the Darpa Urban Challenge-, and the VIAC project \cite{broggi2012autonomous} in Europe, describe similar architectures. The perception, decision, control and actuation stages are the main blocks used.

\begin{figure}[h!]
    \centering
    \includegraphics[width=0.49\textwidth]{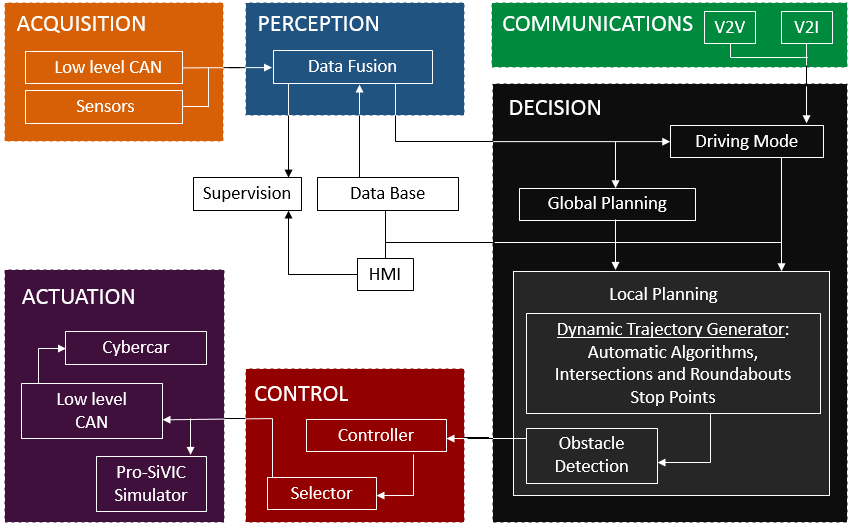}	
    \caption{Control architecture for CTS}
    \label{ControlArchitecture}
 \end{figure}

Recently, RITS team has been improving the path planning and control stage for urban areas \cite{Gonzalez2013} and \cite{Perez2014}. The control architecture for CTSs is shown in Figure \ref{ControlArchitecture}. The main feature is its modular capability, which allows its implementation in simulated (Pro-sivic) and real environments. The software used for the development is \textit{RTMaps}\footnote{http://intempora.com/}. It is a real-time high performance tool, designed for fast and robust implementation in multitask and multivariable systems.

\subsection{Acquisition, Perception and Communications}

The acquisition stage manages the embedded sensors in the vehicle. Using a low level Controller Area Network (CAN) is possible to receive the information describing the environment and the vehicle state (position, speed, accelerations, etc). A detailed description of the available sensors in the vehicle and their requirements is presented in Section \ref{section:requirements}.

The information from the acquisition is processed in the perception stage. Here, fusion algorithms merge the information of the vehicle state (e.g. Kalman filters fusion the SLAM and Inertial Measurement Unit—IMU). Obstacle detection algorithms perceive and classify the different objects and road users in the environment \cite{Trehard2014}.

Communications are provided over WIFI protocol with an Optimized Link State Routing (OLSR) Ad-Hoc network. The transmitted information comes from other vehicles (V2V), providing position, speed, accelerations, planned trajectories, among others. Other static information comes directly from the infrastructure thanks to V2I communications.


\subsection{Decision}

The core of the contributions presented in this work is described in the Decision stage (see Fig. \ref{ControlArchitecture}). Here global and local planning modules are used to generate the path planning and the reference speed per trajectory.

\subsubsection{Global planning} Performs the first planning generation process. A data base module (center part of Figure \ref{ControlArchitecture}) allows to get the information to create a first path, using an XML file. This planner creates a path from the vehicle position to a desired destination. The created path is formed by urban intersection points.

\subsubsection{Local planning} Based on the raw path given by the global planner, a safe and comfortable local trajectory is generated in this module. Based on penalty weight functions, trajectories at intersections, turns and roundabouts can be smoothly planned—as described for simulations in \cite{Perez2014}. Trajectories are computed with B\'ezier curves due to its modularity, fast computation and continuous curvature characteristics; allowing smooth joints between straight and curved segments. 

The trajectory will be set up to the horizon of view, given by the sensors range (approximately 50m). In this sense, an obstacle avoidance module that considers static obstacles have been developed \cite{Perez2014}. This module considers also lateral accelerations, planning the velocity profile according to the level of comfort set by the user (levels described in the \textit{ISO2631-1 Standard} \cite{Labakhua2006}) and the curvature of the path.

\begin{figure}[h!]
    \centering
    \includegraphics[width=0.45\textwidth]{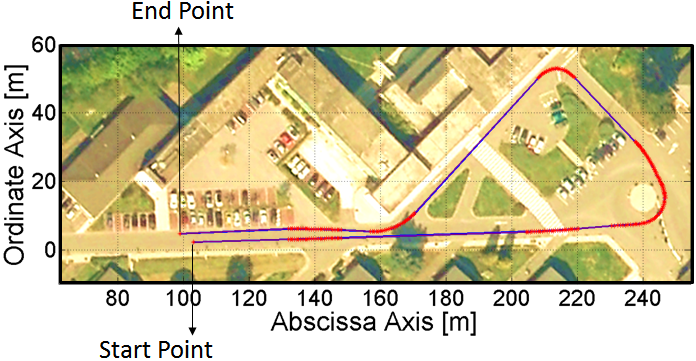}	
    \caption{Local Planning Path at INRIA Rocquencourt}
    \label{fig:LocalPlanning}
 \end{figure}

An example of the path planning generated is shown in Figure \ref{fig:LocalPlanning}. It depicts the performance of the planner at our facilities (handling intersections, turns and roundabouts), were the red points and blue lines describe the curved and straight segments, respectively.

In previous works, only the start and end points of the trajectory were considered in CTSs applications (Fig. \ref{fig:LocalPlanning}). Pre-programed stop points is one of the contributions presented in this work. They are considered in the global and local trajectories (Fig. \ref{ControlArchitecture}).

\begin{figure}[ht]
    \centering
    \includegraphics[width=0.43\textwidth]{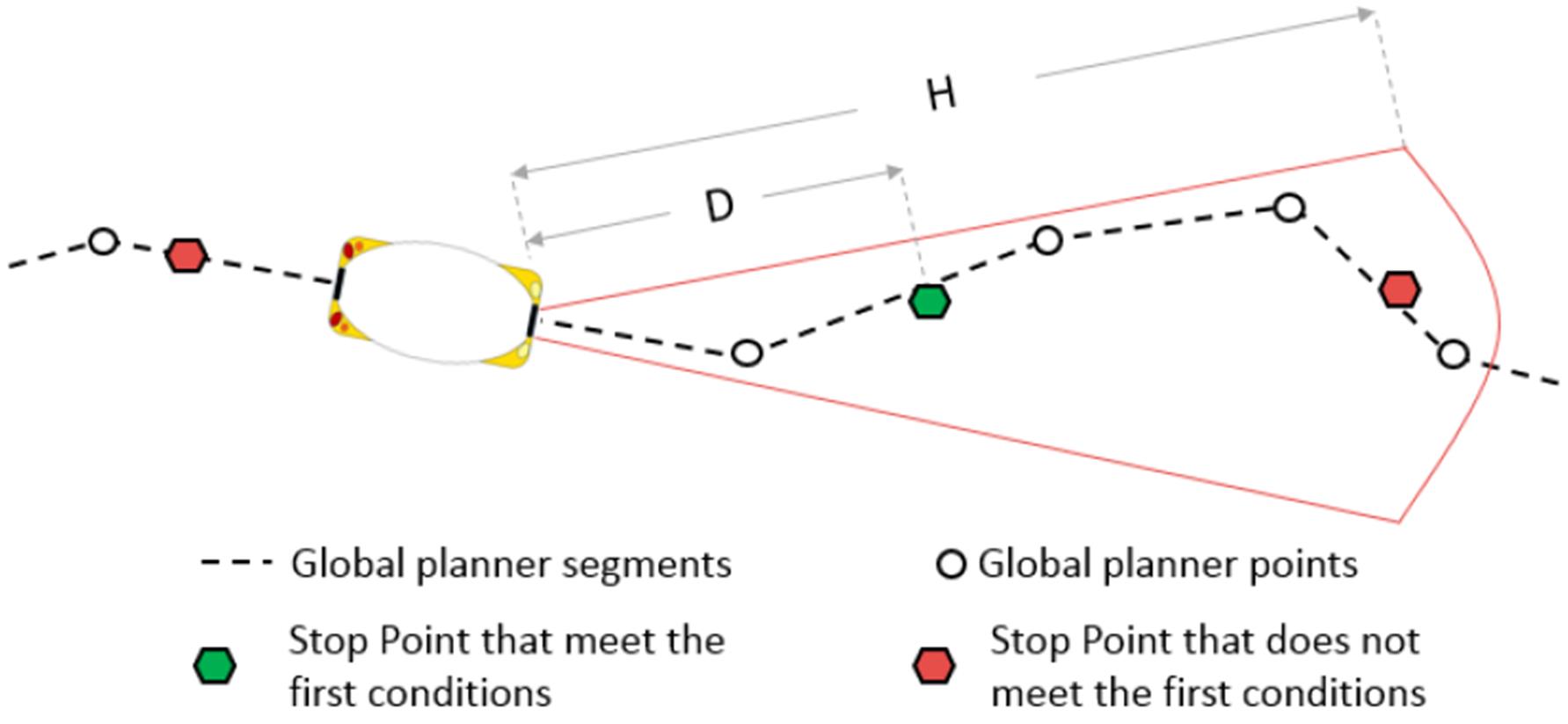}	
    \caption{First set of conditions to consider sending a stop point through the buffer}
    \label{StopPointsFig1}
 \end{figure}

\begin{figure}[ht]
    \centering
    \includegraphics[width=0.40\textwidth]{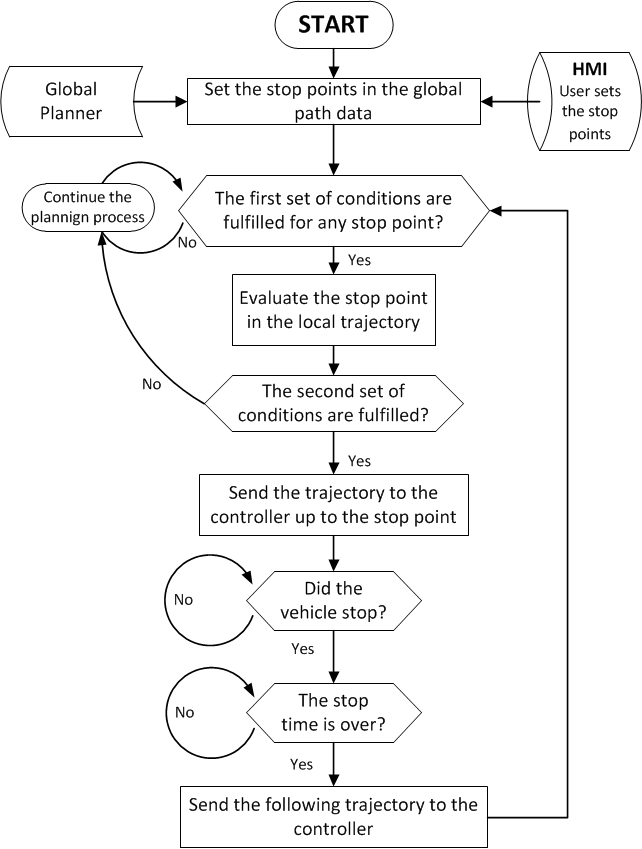}	
    \caption{Stop Points Logical Diagram}
    \label{StopPointsAlgorithm}
\end{figure}

To introduce them in the trajectory, two sets of conditions are evaluated (as showed in Fig. \ref{StopPointsAlgorithm}). The first set evaluates the stop point with respect to the global path. Figure \ref{StopPointsFig1} shows an example of this evaluation, based on the following:

\begin{enumerate}
	\item The stop point is in the current or following segment on the global path (green point in Fig. \ref{StopPointsFig1}).
	\item The distance from the vehicle to the stop point (D) is lower than the horizon of view of the vehicle (H).
	\item The stop point has not yet been sent to the control stage (to avoid considering the same stop point more than once).
\end{enumerate}

\begin{figure}[ht]
    \centering
    \includegraphics[width=0.45\textwidth]{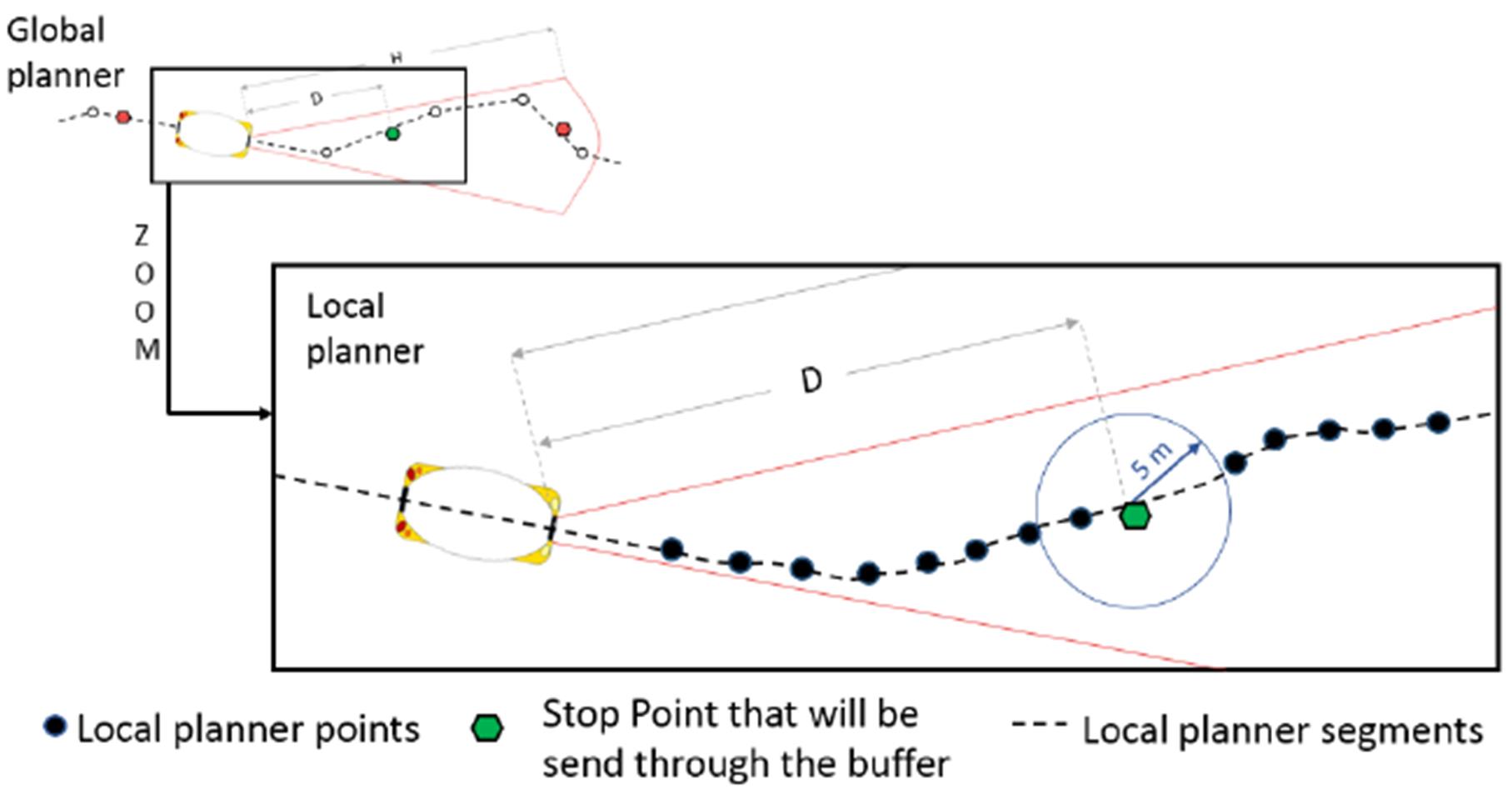}	
    \caption{Second set of conditions that the stop point has to meet to be sent through the buffer}
    \label{StopPointsFig2}
 \end{figure}

\begin{figure}[t!]
	\centering
  \subfigure[]{  
		{\includegraphics[width = 3.05in]{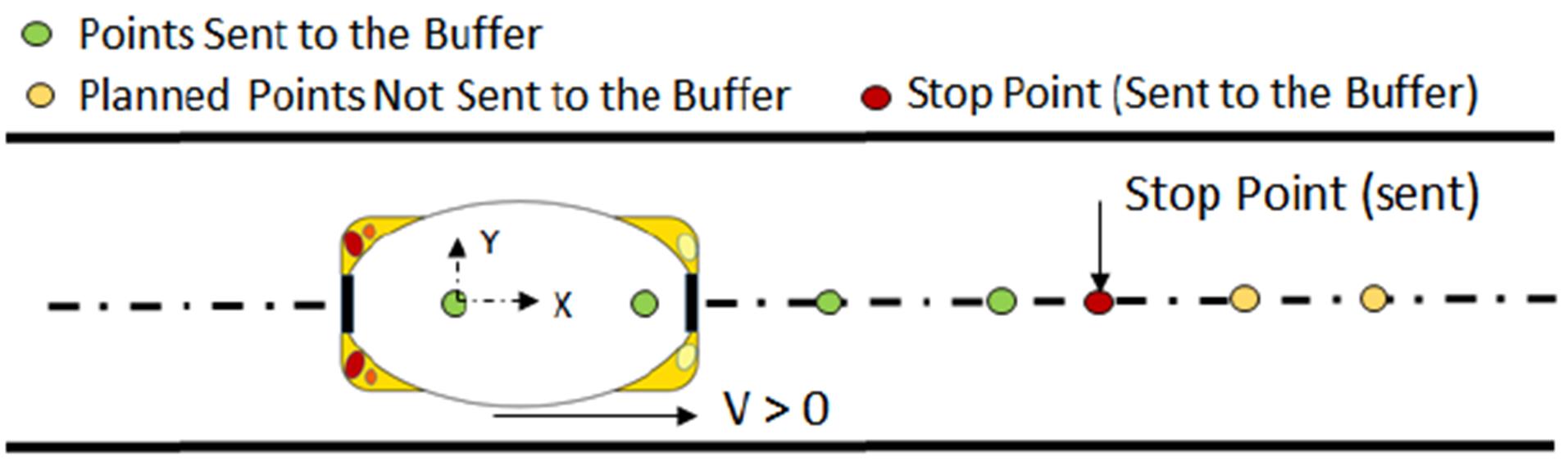}}
		\label{SPSequencyA}}
	\centering
	\subfigure[]{
		{\includegraphics[width = 3.05in]{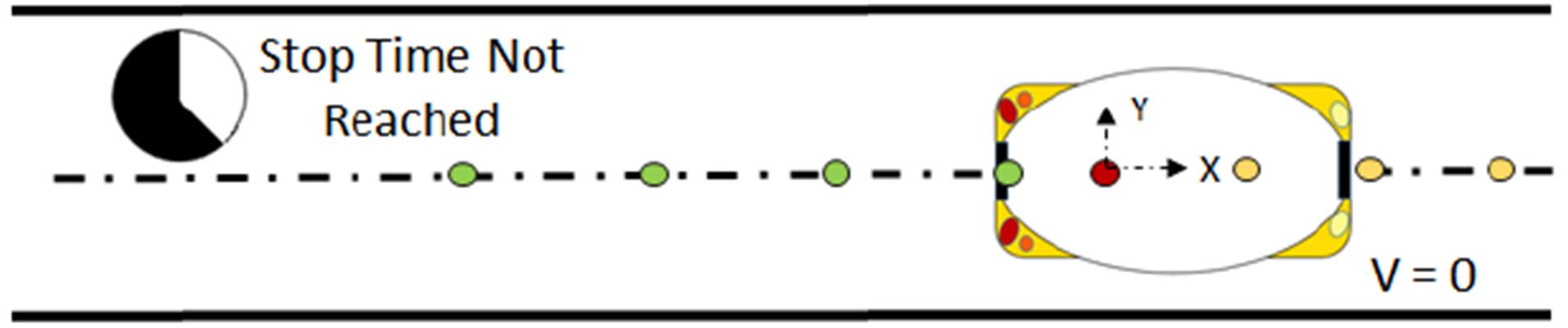}}
		\label{SPSequencyB}}
	\centering
	\subfigure[]{
		{\includegraphics[width = 3.02in]{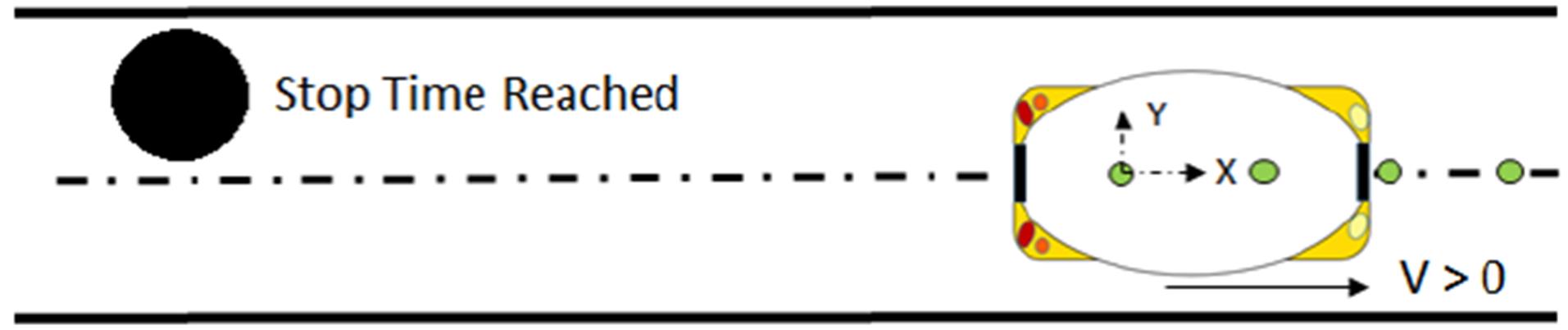}}
		\label{SPSequencyC}}
  \caption{Stop Points and Buffer Behavior}
	\label{SPAlgorithm}
\end{figure}

They remain in a temporary buffer, where a second set of conditions are evaluated regarding the local trajectory (Fig. \ref{StopPointsFig2}). These conditions are:

\begin{enumerate}
	\item The distance between the stop point coordinates and the closer local segment is lower than 5\textit{m}.
	\item The segment where the stop point will be included must not be the last segment in the local planner (as the red point in Fig. \ref{SPSequencyB}).
\end{enumerate}

Then, if the conditions are meet, the position of the stop point is saved into the trajectory with an associated stop time. The trajectory will be sent to the control stage up to the stop point (Fig. \ref{SPSequencyB}). Future trajectory points are not sent to the controller if the desired stop time has not passed (Fig. \ref{SPSequencyC}). 

Based on this method, the CTSs are capable to stop in any part of the trajectory, like curved or straight segments. When the time is reached, the buffer sends the new trajectory, and the vehicle returns to the planned itinerary (Fig. \ref{SPSequencyC})).

\subsection{Control} 

\begin{figure}
    \begin{center}
    \includegraphics[width=0.45\textwidth]{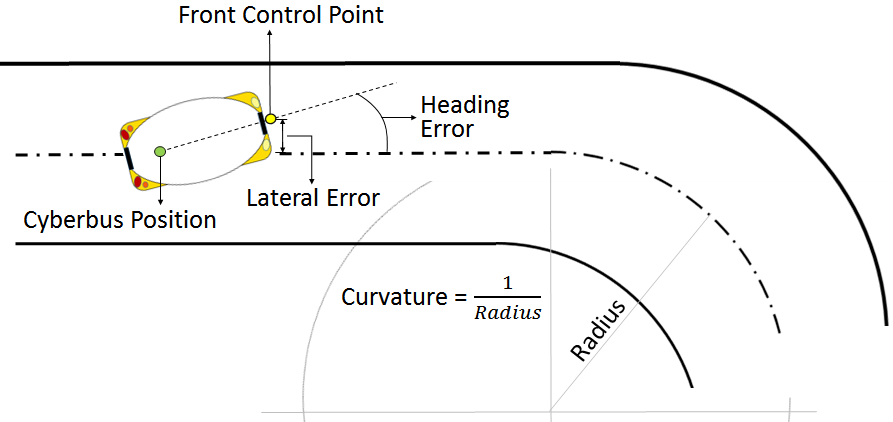}	
    \caption{Lateral Control Variables}
    \label{fig:CybusControl}
		\end{center}
 \end{figure}

The control stage receives the trajectory through the buffer. This feature creates a new trajectory in real time (e.g. lane change, pedestrian detection, etc). The control law implemented is defined in equation \ref{eq:controlequation}:

\begin{equation}
U(t) = {\alpha_1}k(t) + {\alpha_2}L_{error} + {\alpha_3}H_{error}\label{eq:controlequation}
\end{equation}

\noindent where $k(t)$ is the curvature, $L_{error}$ is the lateral error and $H_{error}$ is the heading error (see figure \ref{fig:CybusControl}). The values $\alpha_1$, $\alpha_2$ and $\alpha_3$ are the controller gains.


\subsection{Actuation}

\begin{figure*}[t!]
    \begin{center}
    \includegraphics[width=0.75\textwidth]{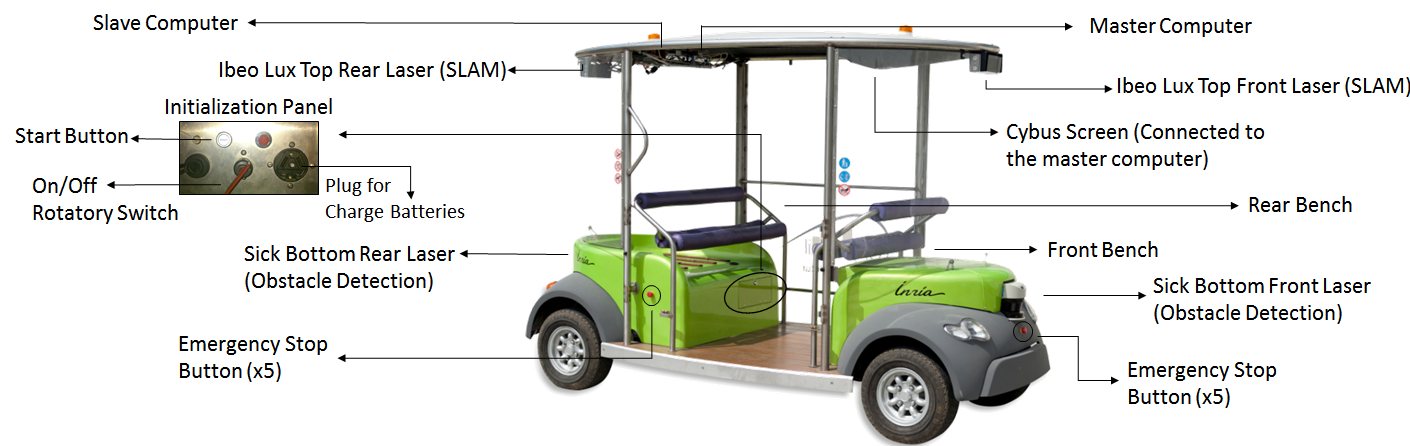}	
    \caption{Hardware devices distribution of Cybus at Inria}
    \label{fig:Cybus}
		\end{center}
\end{figure*}

The actuation stage receives the instructions from the control modules and performs the desired lateral and longitudinal actions on the vehicle. It manages the actuators of the CTS through the CAN protocol, to achieve the desired steering angle and speed. The modularity of the system allows an easy implementation in real platforms, as will be shown in the results section.

\section{CTS equipment at Inria: \textit{The Cyberbus}} \label{section:requirements}

The Cyberbus (Figure \ref{fig:Cybus}) he first is the master, in charge managing the steering, speed, obstacle detection and tracking and control. The slave PC is in charge of executing the localization algorithms, receiving directly the information from the lasers and performing the SLAM.

This prototype CTS was tested during a 3 months demon-stration showed that fully autonomous road vehicles are suitable for public transport \cite{Bouraoui2011}. These demonstrations used previously recorded routes, unable to be modified in real time.

For localization, a Simultaneous Localization and Map-ping (SLAM) algorithm is used. It is capable of estimating the joint posterior probability over all past observations using real time information for a large environment reconstruction, as explained on \cite{Xie2010}. The environment data is collected by two Ibeo Lasers on the top front and rear parts of the Cyberbus (Figure \ref{fig:Cybus}).

Obstacle detection is also possible with the two Sick lasers located on the front and rear down parts of the vehicle. These lasers are capable of detecting an obstacle within a 50\textit{m} range, avoiding possible collisions. The acceleration is tracked by an IMU (located over the rear axle). Two incremental encoders are responsible of measuring the speed (attached to the engines in the rear axle) and a differential encoder tracks the steering angle of the wheels (over the front axle).

The Cyberbus is equipped with two computers. The first is the master, in charge managing the steering, speed, obstacle detection and tracking and control. The slave PC is in charge of executing the localization algorithms, receiving directly the information from the lasers and performing the SLAM. It sends the information to the master PC.

Finally, a joystick and a touchscreen are part of the Human Machine Interface (HMI). All the vehicle is powered by a 72 Volts supply conformed by six Li-Ion batteries. It counts with two DC/DC voltage transformers that reduce the voltage to a 24 volts stage, which supplies the most components of the vehicle. For safety, 6 emergency stop buttons and a Remote Radio Unit (RRU) are capable to suddenly stop the vehicle in case of emergencies.


\section{Tests and results validation} \label{section:results}

This section introduces the experimental validation of the proposed architecture. Several experiments at the INRIA-Rocquencourt facilities were carried out in order to validate the control architecture described, considering different scenarios with multiple stop points. The robustness of the system was tested after numerous automatic performances with two Cyberbus vehicles. Since the experiments were per-formed in real scenarios, some requirements were previously set to guarantee the safety, specifically of the vulnerable road users, based on CityMomil2 regulations \cite{CM22013}. The next subsection describes how these requirements were taken into account.

\subsection{CTS Requirements for Perception and Control}

CTSs can navigate through a whole space with the unique target of reaching the final destination in a safe manner. Functional requirements for both control \cite{CM22013} and perception \cite{PercCM22013} layers were identified within the CityMobil2 project. These requirements were taken into account for the Le´on exhibition. An open square of 40x27 meters where pedestrians are randomly walking in any direction were used as experimental driving area. Pedestrian safety was one of the key requirements when adapting the experimental platform to the driving scenario. The other critical point was the turning radius, that somehow limited the potential areas were the Cyberbus could perform an U-turn. Based on this, the different driving scenarios are depicted in Figure \ref{RoutesDemoLeon}.


\subsection{Experimental Validation}

\begin{figure}[t!]
    \centering
    \includegraphics[width=0.45\textwidth]{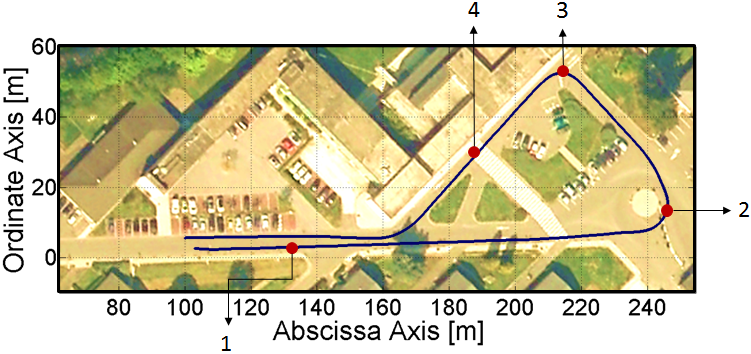}	
    \caption{Stop Points Performance at INRIA-Rocquencourt}
    \label{StopPointsItinerary}
 \end{figure}

The improvements of the local planning, tested in real platforms, are described in this section. Figure \ref{StopPointsItinerary} shows the whole performance of the vehicle in our facilities, based on the path showed in Figure \ref{fig:LocalPlanning}. To validate the stop points algorithm, the same path was performed by the vehicle. However, 4 Stop points were added to the path in this case. Different zones were selected to this end: 2 in straight segments, 1 in a curve and another one inside the roundabout. The complete path is shown in figure \ref{StopPointsItinerary}, the route is described as follows:\\

\begin{itemize}
	\item \textit{The Cybus starts the trajectory at start point (as in figure \ref{fig:LocalPlanning})}
	\item \textit{A stop point is set at $25m$ (first point in figure \ref{StopPointsItinerary}) with a stop time of $30s$.}
	\item \textit{The 2nd Stop Point is set at the roundabout (2nd point in figure \ref{StopPointsItinerary}) with a $25s$ stop time}
	\item \textit{At $45m$, the third stop point is set at a curved segment (3rd point in figure \ref{StopPointsItinerary}) with a stop time of $15s$.}
	\item \textit{The last stop point is set at the straight segment beside the pedestrian crossing, where the vehicle stops for $10s$.}
	\item \textit{Finally, the vehicle finishes successfully the planned trajectory.}
\end{itemize}

\begin{figure}[!ht]
	\centering
  \subfigure[]{  
		{\includegraphics[width = 0.50\textwidth]{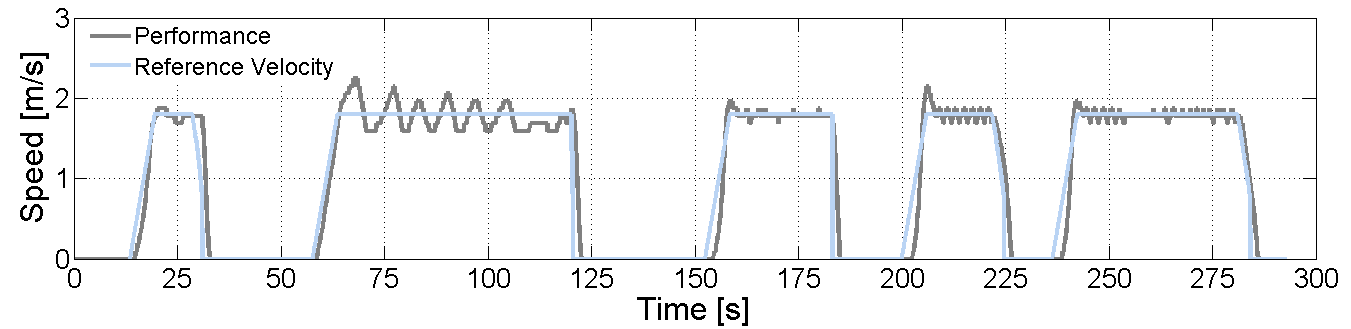}}
		\label{SPRocqGraphicsAA}}
	\centering
	\subfigure[]{
		{\includegraphics[width = 0.50\textwidth]{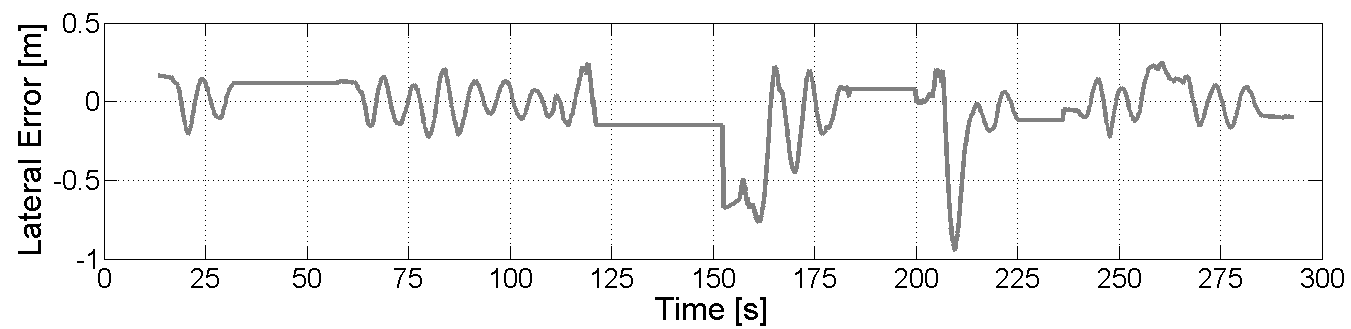}}
		\label{SPRocqGraphicsBB}}
	\centering
	\subfigure[]{
		{\includegraphics[width = 0.50\textwidth]{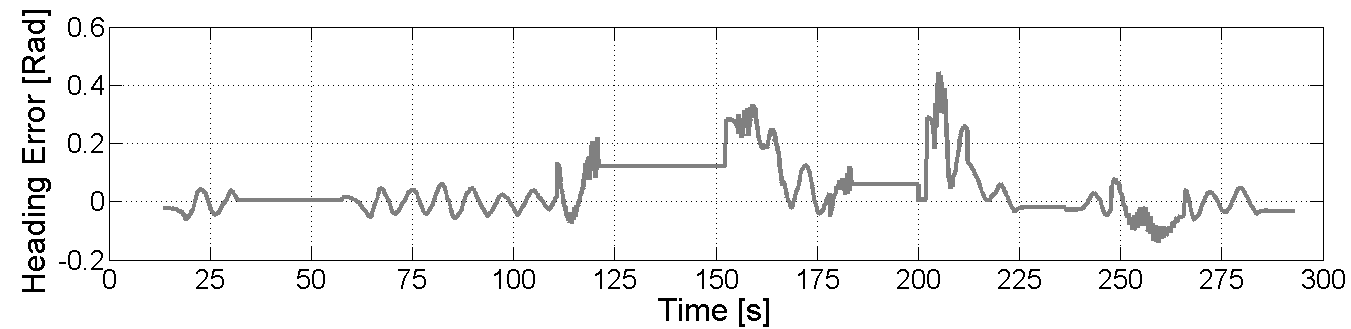}}
		\label{SPRocqGraphicsCC}}
	\subfigure[]{
		{\includegraphics[width = 0.50\textwidth]{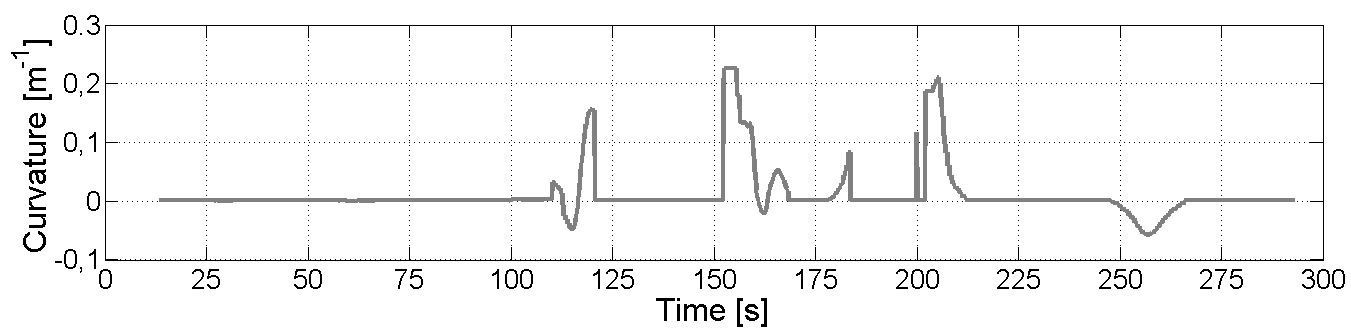}}
		\label{SPRocqGraphicsDD}}	
  \caption{Automatic Test Results at INRIA-Rocquencourt}
	\label{SPRocqGraphics}
\end{figure}

Figure \ref{SPRocqGraphics} shows the performance of the stop points programmed and the trajectory achieved, for the itinerary described in Fig. \ref{StopPointsItinerary}. Figure \ref{SPRocqGraphicsAA} shows the real speed of the vehicle. After the first stop, the actual speed presents some oscillations. It is because this part of the path is a slope-down segment, so the electric engine is braking most of the time. In the other segments, we can appreciate how the CTS reaches the desired velocity smoothly, just reducing the speed when a stop point is close. Figures \ref{SPRocqGraphicsBB} and \ref{SPRocqGraphicsCC} show the two errors used for the lateral control of the vehicle. The curvature is shown in Figure \ref{SPRocqGraphicsDD}. It is important to notice that the errors stay constant when the car stops on a point. Most of time, the errors are close to zero. The lateral error average in curve segment is less than 0.5m and the heading error is between -0.2 and 0.2 rad. Finally, the curvature never reaches the steering limit of the car, i.e. 0.48 [1/\textit{m}].

\subsection{Showcase demonstration at Le\'on}

\begin{figure}[ht!]
    \centering
    \includegraphics[width=0.45\textwidth]{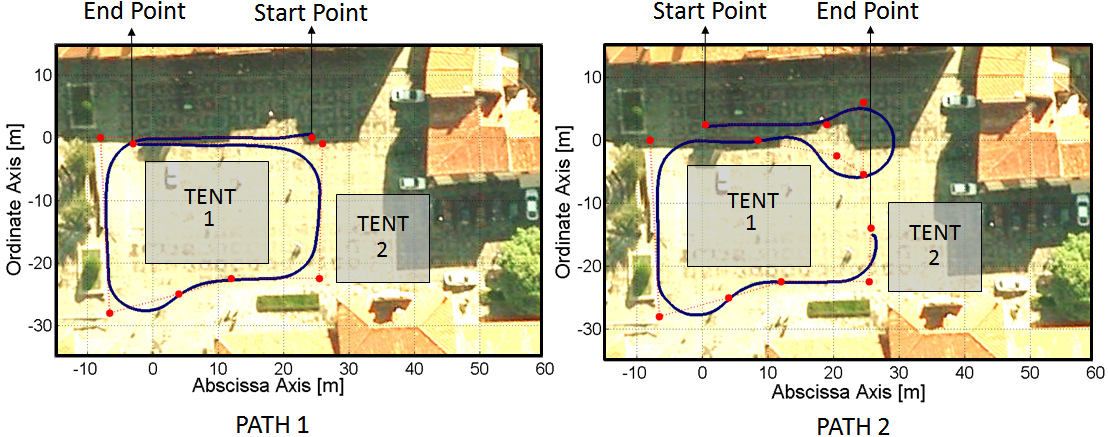}	
    \caption{Different Driving Scenarios at Le\'on Exhibition}
    \label{RoutesDemoLeon}
 \end{figure}

Results obtained in the first showcase of the CityMobil2 project are presented in this section. Figure \ref{RoutesDemoLeon} shows two trajectories performed in a public open square. Two tents were located in the middle square for expositions related to the project, and the traffic on the next street was regulated during the demonstration. The space was reduced, for this reason a path around the big tent (left-hand side of Figure \ref{RoutesDemoLeon}), and other performing an U-turn (right-hand side of the same figure), were carried out. The first scenarios shows how the global control points were defined (as explained in Section II), then the blue line shows the performance of the vehicle in autonomous driving.

\begin{figure}[!ht]
	\centering
  \subfigure[]{  
		{\includegraphics[width = 0.50\textwidth]{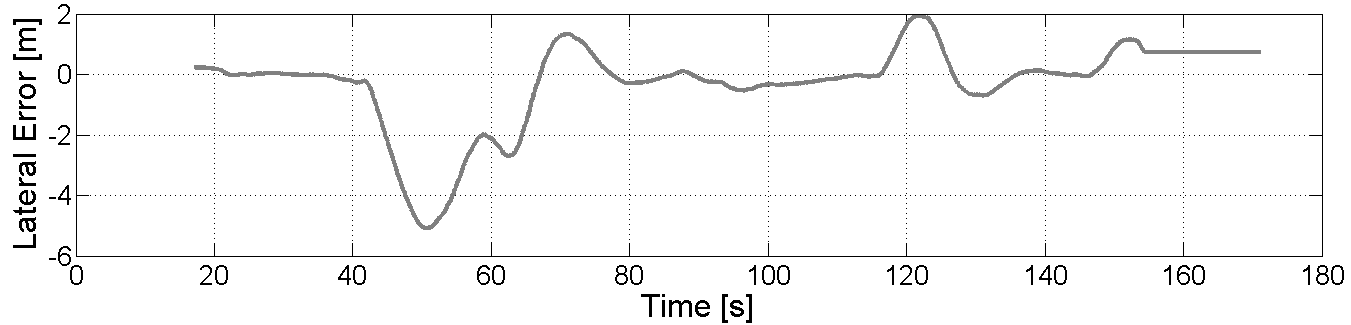}}
		\label{LeonExhibitionAA}}
	\centering
	\subfigure[]{
		{\includegraphics[width = 0.50\textwidth]{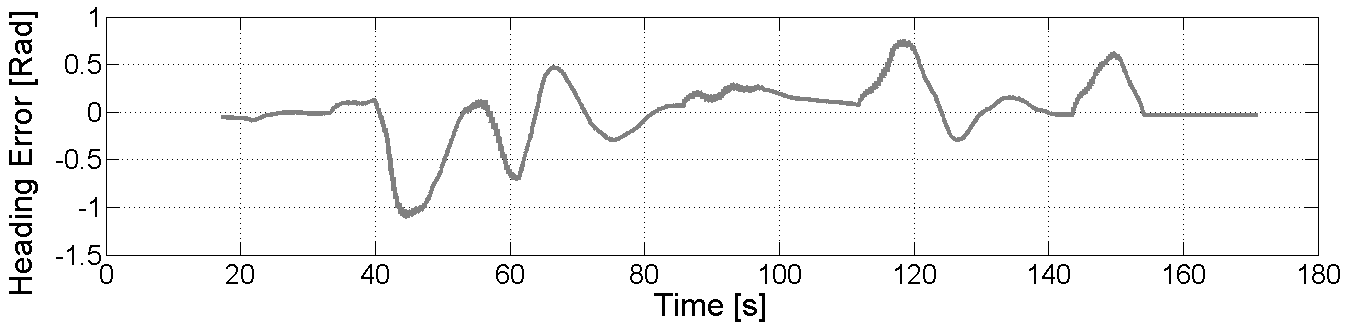}}
		\label{LeonExhibitionBB}}
	\centering
	\subfigure[]{
		{\includegraphics[width = 0.50\textwidth]{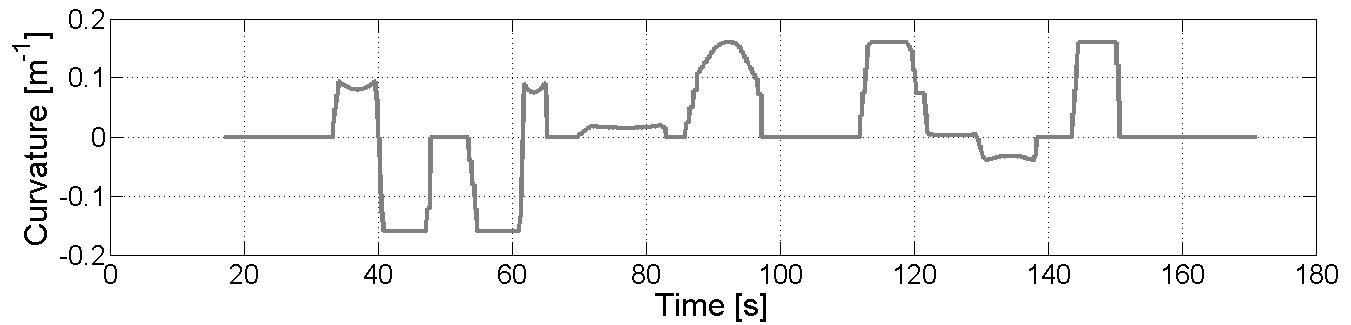}}
		\label{LeonExhibitionCC}}
	\caption{Vehicle Performance at Le\'on Exhibition}
	\label{LeonExhibition}
\end{figure}

Figure \ref{LeonExhibition} shows the vehicle performance. The lateral and heading remains small in straight segments, and only exceeds the average of the first experiments around the 60 seconds of the experiment. It is because the angles at this curve (The 3rd one, in the bottom right side of the Figure \ref{RoutesDemoLeon}) is more pronounced that the other, and also at this time the vehicle is descending the sidewalk. But, the oscillation is reduced by the control stage. Figure \ref{RoutesDemoLeon} shows another trajectory. In this case, due the space limitations, the vehicle achieves the U-turn in around 13 meters. The maximum curvature radius of the CTS is reached, and the controller stabilizes the system when vehicle exits the curve.

\section{Conclusions}

In this paper, the control architecture for CTSs has been described in the framework of the FP7 project CityMobil2. Stop points were introduced and tested in real scenarios. The key stop points throughout the circuit of the vehicle are set, improving the performance of the autonomous CTSs. These are an added special feature of the Local Planning.

Two specific experiments were performed: a demonstra-tion in our facilities and a CityMobil2 showcase at Le\'on (Spain). The results of these experiences are described, showing good results, where the performance of the vehicle is described in terms of lateral and heading errors, as well as the curvature and longitudinal speed. The acceptability of this technology is one the objective of the project. In this demonstration more than 50 persons tested our vehicles, with good impressions.

Future works will consider the introduction of vehicle-to-vehicle communications among CTSs in order to intelligently handle complex traffic situations as intersections or round-abouts.


\section*{ACKNOWLEDGMENT}

Authors wants to thank to the ARTEMIS project DESERVE and FP7 CityMobil2 for its support in the development of this work.


\bibliographystyle{IEEEtran}
\bibliography{ICVES2015}

\end{document}